YIGIT OKTAR

Izmir University of Economics

June 3rd 2016

# Depth Estimation from Single Image using Sparse Representations

Technical Report

**Introduction**

Monocular depth estimation is an interesting and challenging problem as there is no analytic mapping known between an intensity image and its depth map. Recently there has been a lot of data accumulated through depth-sensing cameras, in parallel to that researchers started to tackle this task using various learning algorithms. [1] [2] [3] [4]

**Previous Work**

One of the first works in literature that achieved satisfactory results in monocular depth prediction was based on Markov Random Fields. [1] MRF can be described as an undirected graph composed of random variables satisfying Markov property, which ensures future states to depend only on the current state. MRFs are extensively used in probabilistic image modeling. In the study noted, certain local and global features were used to model relative and absolute depths at pixels. This model was then used to infer depth maps for test images.

In a more recent study, depth estimation was formulated as a minimization problem as a combination of loss functions, main loss function being depth transfer. [2] This function enforces correspondences in appearance, where multi-scale SIFT features were used as appearance cues. In this approach, predicted depth map optimizes appearance correspondence, favors smoothness, and abides by global probability distribution all at the same time, in a non-parametric way.

Within last two years, deep learning approach has resulted in many successful applications. In a recent study, in which a combination of two deep networks was utilized to achieve great results in depth estimation. [3] In this approach, initial convolutional neural network predicts depth globally, and feeds its input to another, which further refines prediction locally.

In a more recent deep learning approach, convolutional neural networks were integrated into a continuous conditional random field. [4] In this approach, again there are two convolutional neural networks, one for local and one for global predictions. This time, there is a final analytical layer (CRF loss layer), which outputs a final depth map by considering both local and global predictions and enforcing continuity in a probabilistic manner.

**Proposed Method for Dictionary Learning**

Dictionaries learnt using sparse representations can be applied on classification tasks. It is important to note that, $L_1$(Basis-Pursuit) is shown to be equivalent to an SVM. [5] SVMs were the best machine learning tools, before deep learning bloomed. $L_0$ is perhaps still powerful, as an NP-hard construction. However, CNNs dominate the literature in all image related learning tasks and have achieved state-of-the-art results in many. We propose a method to learn dictionaries by deep sparse coding, and compare its performance with previous studies mentioned earlier.

$$A_1 X_1 = Y$$
$$A_2 X_2 = X_1$$
$$A_3 X_3 = X_2$$
$$....$$

$$\Rightarrow A_1 A_2 A_3 .... A_n X_n = Y$$

In this method, the signal is first used to learn a dictionary via a conventional dictionary learning method. Then, the resulting sparse representation is passed to successive layer as an input, on which the second dictionary learning is performed. Iterating in this manner as much as desired, the resulting sparse code can be regarded as a "deep" one. Note that, one can recover the full dictionary by multiplying dictionaries of each layer. Such method will have some advantages and also difficulties as listed below:

**Advantages:**
- We can depict dictionaries at different layers and may observe meaningful relationships between dictionary atoms and the nature of the signal. Analyzing layers, we might get insight on its properties.
- Such layered structure can provide flexibility. For example, different metrics can be applied at different layers, depending on the problem at hand.
- A non-linear function can be applied on the coding of an intermediate layer.
- There is a faster convergence possibility if a proper design is used.

**Difficulties:**
- If design is not theoretically sound, too much error will build up while descending layers, and convergence will not be guaranteed.
- It requires a design that will always guarantee convergence. An equivalent formulation of backpropagation in neural networks should be considered.
- Dictionary initialization becomes problematic if random methods are used. Randomness in each layer will drastically effect overall efficiency.

**Dictionary Initialization in a Deep Model**

Random initialization of dictionaries is a bit problematic. First of all, it is not an analytic approach, so good results are not consistently guaranteed. Yet another downfall of random approach is that, in order to recreate the results, we need to save the seed for the generator. An improved version of fully random dictionary is to start with a dictionary consisting of a random subset of input data patches. However, again it may fail as it depends on randomly chosen subset. Moreover, random dictionary initialization will be more problematic for deep structures, as randomness in each layer will drastically effect overall efficiency, and will result in a very low convergence rate.

First, we depict the comparison table for some constant dictionaries used on Barbara. These were used for sparse coding + approximation with OMP. For all results parameters were: patch size: 4x4, sparsity=2, #atoms = 64

|  | Uniform | Random | Random Perm. | DCT |
|---|---|---|---|---|
| PSNR Value | ~11.5 | ~19.5 | ~26.5 | ~26.7 |

DCT seems to perform best in this case. Note that, these were constant dictionaries used in sparse approximation without any dictionary update steps.

Shown below is the table for PSNR values after some iterations of sparse coding + dictionary updates, taking each of dictionaries above as starting dictionaries. Note that, deep sparse coding model (3-layered) was used, 1st layer dictionary initialized with these listed ones.

|  | Uniform | Random | Random Perm. | DCT |
|---|---|---|---|---|
| PSNR Value | ~24 | ~29 | ~30 | ~27.5 |

Here we observe that DCT no longer performs best in a deep structure. As a constant dictionary, it is good as a general one, but it is perhaps not really adaptable as its improvement is not very striking. On the other hand, random and random permutation starters perform really well. This is probably because of their inherent flexibility. Note that, we can get low PSNR values in some cases of random and random permutation starters as they are not deterministic.

**Initialization with Best Patches**

We propose a deterministic dictionary initialization using a subset of data patches. Instead of choosing a randomly permuted subset of patches, we choose a subset of patches that approximates all of the patches well. This can be accomplished by sparse coding itself.

In this method, all data patches are passed to the algorithm along with column size K of the dictionary to be created. A temporary dictionary A is initialized consisting of all patches. Namely, YX = Y. Then sparse coding of X is performed with sparsity 1. The rows of resulting X, with higher l-0 norms, will approximate the whole data better, as each row of X corresponds to an atom, which is actually a patch from the data. A higher l-0 norm means, corresponding patch is a dominant signal, and will be a better option than a randomly chosen one. As a final step, rows of X can be sorted by their l-0 norms in descending order, acquiring first K indices. These indices will be the indices of patches that will initialize the dictionary desired.

With this method we were able to surpass random permutation method for initializing a deep structure, acquiring a PSNR value around 32, as opposed to 30 as listed earlier.

In a deep model, inner dictionaries could be initialized with this method and probably it will improve results. However, as this method increases time complexity, a complete evaluation was not performed on this issue.

Performance of initializing dictionaries with data patches depends on the nature of data at hand. For certain tasks, this method might not be the most effective. Before deciding on whether to use it or not, analyzing problem's requirements and properties of feature space will be beneficial. For example, some problems require a more flexible dictionary initialization step which can easily be achieved by randomness. Therefore, feature extraction from raw data is as important as dictionary initialization.

**Feature Extraction from Depth Data**

NYU v2 depth dataset was processed using two feature extraction variations. In first variation, 8x8 patches of intensity values, that map to a block of constant depth values, were extracted. In the second variation, instead of constant depth values, mode of depth block was used as the label for that patch. For each depth (from 1 to 10), 10 distinct dictionaries were trained using deep sparse coding method proposed.

As variations in deep model, 2-layered and 3-layered structures were tested. In 2-

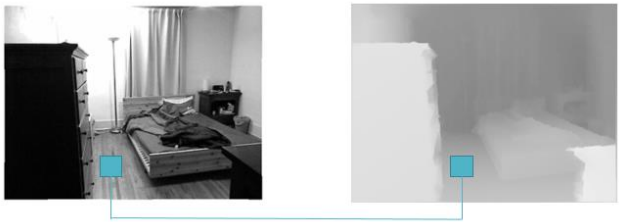

layered structure, first layer was fixed with DCT, whereas in 3-layered one, all dictionaries were randomly initialized. Sliding window was also tried in deeper layers, to mimic convolutional neural network logic. However, since computational complexity was high for such cases, no reasonable evaluation was possible.

In 2-layered model, in which first layer was fixed with DCT, dictionary formed in $2^{nd}$ layer has some interesting properties. Nearly all atoms that are apparent have high frequencies, more specifically, they have checkerboard appearance with certain parts suppressed. It is possible that high frequency formulations are significant in depth processing.

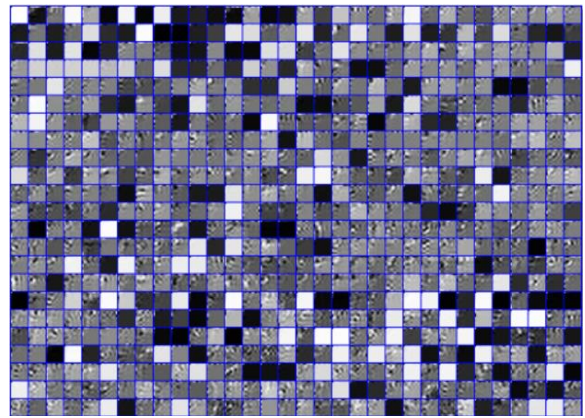

$2^{nd}$ layer dictionary of 2-layered model

Dictionaries formed in 3-layered model had a lot of repetitive patterns. Also, there was no striking distinction between dictionaries for near and far. This was probably caused by initializing all layers randomly, which resulted in a very low level of convergence. Therefore, those dictionaries are omitted.

## Classification Scheme

Two different schemes were tested. In one, all distinct dictionaries were concatenated into one, then depth-d dictionary, with most number of non-zero coefficients in the code, was chosen to label that patch as depth d.

In a different scheme, dictionaries were used distinctly and then dictionary resulting in a sparse code with least l-1 norm was chosen to label that patch. Both gave similar results.

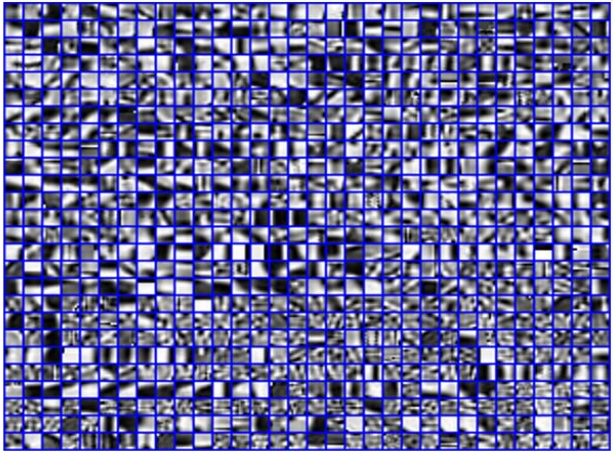 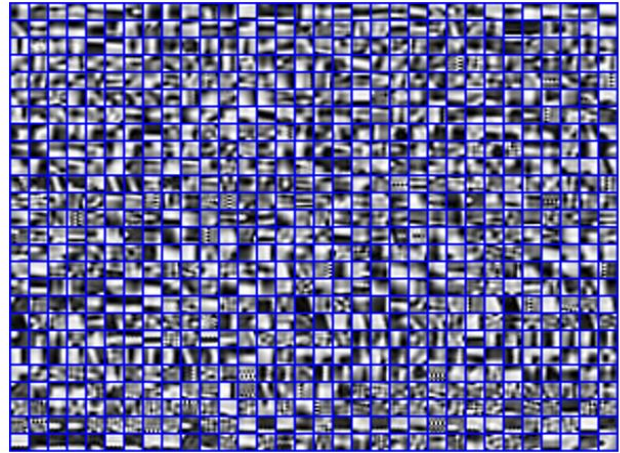

**Near** **Far**

Full dictionaries in 2-layered model, for depth 1 and depth 10 respectively.

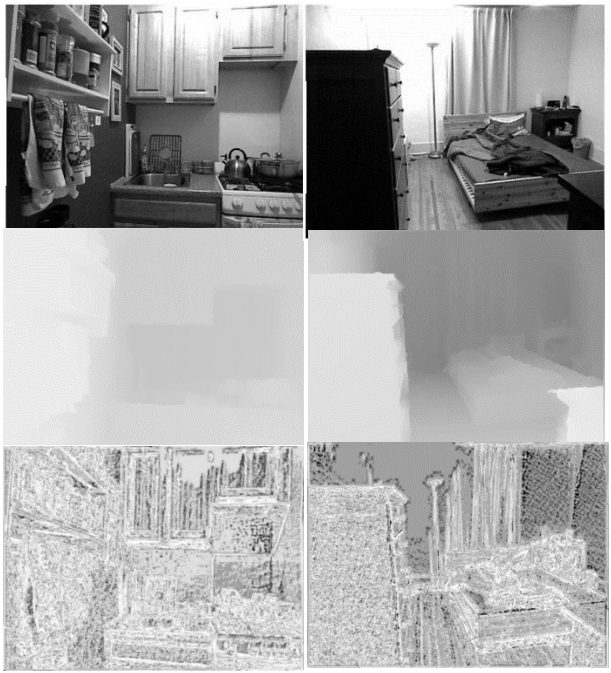

Predictions for #250 and #540 using 2-layered model, resulting in PSNR values of 17.21 and 18.65

| Method | Avg. PSNR |
|---|---|
| Deep Sparse (3-layered) | 15 |
| Deep Sparse (2-layered) | 17 |
| Depth Transfer (Karsch, 2012) | 20 |
| Deep Networks (D. Eigen, 2014) | 22 |
| Conv. Neural Fields (F. Liu, 2014) | 25 |

### Results

We present our results for two images taken from NYU v2 dataset and also compare our results (in average) to the methods mentioned previously. Our results are not satisfactory. However, they are still promising. A sense of depth is perceivable in the results. Both deep sparse models performed better than random (PSNR 7) and uniform (PSNR 10) prediction.

Mode of depth block as feature extraction was also tested. However, it did not perform as well as constant depth block extraction.

### Problems in Our Approach

The most severe deficiency is performing only local prediction. There is no global enforcement applied, in our approach, whereas all other previous works have considered it. Also, similarly, there is no preference for continuity of predicted depth field in our approach, which is observable in our results. Our predictions look very discretized even though some Gaussian smoothing is applied.

Other important factor is the criteria for classification. Classification phase is crucial. Even if you learnt conforming dictionaries, if you cannot utilize them through a correct classification scheme, then accuracy is doomed.

As a final note, learning distinct dictionaries may not be the best approach. Classes can be integrated into deep layer structure, where last layer maps input to a class. However, this approach must be based on a theoretical formulation to be successful.

### Future Work

Rather than taking an experimental path, a theoretical framework has to be established on deep sparse coding. It will pave the way to an investigation of parallels between other deep learning approaches. At this point, we cannot say whether one can prove an equivalence between convolutional neural networks and deep sparse representations utilizing sliding window; however, working towards such a proof will be beneficial for sparse coding theory in general.

A recent study claims that they outperformed K-SVD using a greedy deep dictionary learning algorithm. [6] It will be beneficial to study that approach in detail and build on top of that if possible.

### Conclusion

We tried to apply a deep learning strategy on sparse representations for the problem of monocular depth estimation. A relatively easier problem can be targeted for initial experimental investigations. Although, our experimental method did not meet our initial expectations, it is possible that a theoretical approach will be more conclusive whether a deep sparse coding approach can reach state-of-the-art performance in image related tasks.